\documentclass[pdflatex,sn-mathphys-num]{sn-jnl}

\usepackage{amsmath,amssymb,amsfonts}%
\usepackage{amsthm}%
\usepackage{mathrsfs}%
\usepackage[title]{appendix}%
\usepackage{xcolor}%
\usepackage{textcomp}%
\usepackage{manyfoot}%
\usepackage{algorithmicx}%
\usepackage{listings}%
\usepackage{algorithm}
\usepackage{algpseudocode}
\usepackage{adjustbox}
\usepackage{multirow}   
\usepackage{multicol}   
\usepackage{booktabs}   
\usepackage{array}      
\usepackage{graphicx}   
\usepackage{multirow,adjustbox,booktabs,threeparttable,array}

\theoremstyle{thmstyleone}%
\theoremstyle{thmstyletwo}%

\theoremstyle{thmstylethree}%

\begin{document}

\title[Article Title]{MetaQAP - A Meta-Learning Approach for Quality-Aware Pretraining in Image Quality Assessment}


\author*[1]{\fnm{Nisar} \sur{Ahmed}}\email{nisarahmedrana@yahoo.com}

\author[2]{\fnm{Gulshan} \sur{Saleem}}\email{gulshnsaleem26@gmail.com}

\author[3]{\fnm{Nazik} \sur{Alturki}}\email{namalturki@pnu.edu.sa}

\author[4]{\fnm{Nada} \sur{Alasbali}}\email{nalasbali@kku.edu.sa}

\affil*[1]{\orgdiv{Department of Informatics and Systems}, \orgname{University of Management and Technology}, \orgaddress{\city{Lahore}, \postcode{54000}, \state{Punjab}, \country{Pakistan.}}}

\affil[2]{\orgdiv{Faculty of Information Technology and Computer Science}, \orgname{University of Central Punjab}, \orgaddress{\city{Lahore}, \postcode{54000}, \state{Punjab}, \country{Pakistan.}}}

\affil[3]{\orgdiv{Department of Information Systems}, \orgname{College of Computer and Information Sciences, Princess Nourah bint Abdulrahman University}, \orgaddress{\street{P.O. Box 84428}, \city{Riyadh}, \postcode{11671}, \country{Saudi Arabia}}}

\affil[4]{\orgdiv{Department of Informatics and Computer Systems}, \orgname{College of Computer Science, 
King Khalid University}, \orgaddress{\city{Abha}, \postcode{61421}, \country{Saudi Arabia}}}

\abstract{Image Quality Assessment (IQA) is a critical task in a wide range of applications but remains challenging due to the subjective nature of human perception and the complexity of real-world image distortions. This study proposes MetaQAP, a novel no-reference IQA model designed to address these challenges by leveraging quality-aware pre-training and meta-learning. The model performs three key contributions: pre-training Convolutional Neural Networks (CNNs) on a quality-aware dataset, implementing a quality-aware loss function to optimize predictions, and integrating a meta-learner to form an ensemble model that effectively combines predictions from multiple base models. Experimental evaluations were conducted on three benchmark datasets: LiveCD, KonIQ-10K, and BIQ2021. The proposed MetaQAP model achieved exceptional performance with Pearson Linear Correlation Coefficient (PLCC) and Spearman Rank Order Correlation Coefficient (SROCC) scores of 0.9885/0.9812 on LiveCD, 0.9702/0.9658 on KonIQ-10K, and 0.884/0.8765 on BIQ2021, outperforming existing IQA methods. Cross-dataset evaluations further demonstrated the generalizability of the model, with PLCC and SROCC scores ranging from 0.6721 to 0.8023 and 0.6515 to 0.7805, respectively, across diverse datasets. The ablation study confirmed the significance of each model component, revealing substantial performance degradation when critical elements such as the meta-learner or quality-aware loss function were omitted. MetaQAP not only addresses the complexities of authentic distortions but also establishes a robust and generalizable framework for practical IQA applications. By advancing the state-of-the-art in no-reference IQA, this research provides valuable insights and methodologies for future improvements and extensions in the field.}




\maketitle

\section{Introduction}
\label{sec:introduction}
The Human Visual System (HVS) is an intricate and highly evolved sensory system responsible for our ability to perceive the world visually. Visual perception is crucial, with over seventy percent of information intake occurring through vision. The visual cortex in the human brain processes these visual stimuli, making our visual system the most sophisticated among all species. Consequently, visuals have become the primary source of information for humans. However, the quality of these visuals is highly sensitive to both the intrinsic characteristics of the image and the observer’s visual perception capabilities. With the rise of digital multimedia technologies, the importance of visuals in communication and information presentation has surged \cite{li2023quality}.\\
Digital images, captured by digital cameras, are electronic representations of objects. These images can suffer from quality degradation due to artifacts introduced during acquisition, processing, storage, or transmission \cite{ahmed2022biq2021}. Therefore, it is essential to evaluate the performance of systems handling these images by assessing their quality. Perceptual Image Quality, referring to how humans perceive the quality of an image, is crucial since artifacts lead to data loss.\\
IQA can be subjective or objective. Subjective quality evaluation, where humans assess the image, is the most reliable but impractical for most scenarios due to resource constraints \cite{ahmed2022biq2021}. Thus, objective methods using algorithms are employed, making IQA applicable to various fields like broadcasting, video streaming \cite{aslam2024tqp}, image processing enhancement \cite{ahmed2022image}, and real-time quality assessment \cite{saleem2024edge}. Traditional objective methods, such as Mean Squared Error (MSE) and Peak Signal-to-Noise Ratio (PSNR), rely on reference images for quality assessment but often correlate poorly with human perception \cite{khalid2021gaussian,ahmed2021piqi}. More advanced methods like the Structural Similarity Index Metric (SSIM) \cite{wang2004image} and its variants (MS-SSIM \cite{wang2003multiscale}, 3-SSIM \cite{li2009three}, CW-SSIM \cite{sampat2009complex}, IW-SSIM \cite{wang2010information}) show better performance but still require reference images, limiting their practical application.\\
No-reference IQA methods play a critical role in various practical applications where reference images are not available. However, most research in this area often relies on small datasets that offer limited diversity in types of image distortions \cite{ahmed2022biq2021,hosu2020koniq}. This limitation leads to significant drops in performance when algorithms trained on one type of distortion are applied to other types of distortions. Moreover, algorithms trained on artificially simulated distortions tend to struggle when faced with real-world images that contain multiple, authentic distortions simultaneously \cite{song2022blind,ahmed2022biq2021}. Compared to full-reference IQA methods that have the advantage of a clear reference image, no-reference IQA methods face challenges in accurately assessing image quality. Despite efforts to improve by training regression algorithms using real-world statistical features, these methods still encounter difficulties in achieving robust and generalized performance across diverse image conditions \cite{aslam2023vrl}.\\
The necessity for more accurate and generalizable IQA methods is evident, given the vast variations in content and distortion types that small datasets cannot capture \cite{ahmed2022biq2021,hosu2020koniq}. This study addresses this by training deep learning models on large, representative datasets with authentic distortions. Specifically, we focus on CNNs, which have shown remarkable performance in various visual recognition tasks. The objective of the study is to achieve highest correlation with subjective quality evaluation. Following are the specific contributions of this study:
\begin{itemize}
    \item Developed a method to pre-train CNN models on a quality-aware dataset to enhance the model's understanding of image quality.
    \item Designed a loss function specifically tailored for image quality assessment, improving the model's predictive performance.
    \item Utilized meta-learning to create an optimized ensemble model, combining multiple base models for superior performance.
    \item Performed comprehensive experiments, including baseline assessments, cross-dataset evaluations, and ablation studies, to validate the proposed approach.
    \item Demonstrated the model's ability to generalize across different datasets, ensuring robust performance in various real-world scenarios.
\end{itemize}
\subsection*{Paper Organization}
The rest of the paper is organized as follows:  An overview of related work is provided in Section 2 along with a review of current image quality assessment approaches.  Section 3 covers the proposed approach along with the dataset requirements, training methodology and model architecture.  Section 4 provides the experimental setup and evaluation criteria. Section 5 provides the experimental results to validate the performance of the proposed model.  Section 6 concludes the paper with key findings and recommendations for subsequent investigations.
\section{Literature Review}
No-reference IQA is the most challenging type of IQA, attracting increasing attention from the research community \cite{chandler2013seven}. Distortion-specific quality assessment methods cater to particular categories of image distortions, making them suitable for specific applications \cite{yue2023perceptual,li2023agiqa,wang2022generation,zhang2020data}. This study, however, targets general-purpose IQA, which does not focus on any specific distortion category, thereby adding complexity to the problem. The literature review first presents conventional approaches \cite{khalid2021gaussian,ahmed2021piqi,ahmed2020image} based on natural scene statistics (NSS), followed by deep learning-based solutions and recent advances in IQA, including the integration of various models and attention mechanisms.\\
Traditionally, NSS has been employed to address the problem of no-reference and full-reference IQA \cite{chandler2013seven}. Techniques based on NSS leverage statistical properties of natural images to assess quality without reference images \cite{khalid2021gaussian,ahmed2021piqi}. However, with the advent of deep learning, which has outperformed NSS in image recognition, the focus of IQA research has shifted towards deep learning approaches \cite{aslam2023vrl,chandler2013seven,ahmed2022deep}. Deep learning features, learned automatically from training images, eliminate the need for manual and error-prone feature engineering \cite{ahmed2022deep}. Despite the challenge of limited training data, researchers have developed deep learning-based solutions for IQA, addressing the issue of insufficient training images by leveraging pre-trained models and transfer learning \cite{chandler2013seven}. These advances have significantly improved the performance and robustness of no-reference IQA models, making them more aligned with human perception of image quality.
\subsection*{Deep Learning based IQA Approaches}
Deep learning-based approaches have revolutionized the field of IQA, offering significant advancements over traditional methods \cite{aslam2023vrl}. By leveraging the power of CNNs and other deep learning architectures, these methods automatically learn and extract features from training images, thereby eliminating the need for manual feature engineering. Recent studies \cite{aslam2023vrl,ahmed2022deep, aslam2024tqp, ahmed2023perceptual} have demonstrated various deep learning architectures tailored for IQA, each addressing unique challenges and improving upon traditional methods.\\
One notable approach is the use of CNNs for IQA, where the model predicts image quality without any reference image. For instance, Bosse et al. \cite{bosse2017deep} proposed a CNN-based model that learns from human-annotated datasets to predict quality scores, achieving state-of-the-art performance on several benchmarks. Similarly, Gao et al. \cite{gao2018blind} used the Vgg16 architecture for quality evaluation. The extracted features from various layers of Vgg16 and reasoned that different convolution layers reflect image quality differently. They trained an SVR on features acquired at various convolution stages and performed average pooling to produce quality score.\\
Two-stage frameworks emerged as another alternative in which the quality assessment is performed as part of multi-task learning. Zhang et al. \cite{zhang2020blind} presented a two-step framework where the first stage detects the type and degree of distortion, and the second stage uses a specific CNN model to assess image quality. Their models were developed to make quality judgments on images distorted both legitimately and artificially. Kim et al. \cite{kim2018deep} proposed a two-stage approach to predict image quality: the first stage predicts a subjective quality score, and the second stage predicts an objective error map. Similarly, Ravela et al. \cite{ravela2019no} addressed the problem in a two-stage fashion, where the first stage predicts the type of distortion, and the second stage uses a deep model to predict the subjective score, followed by average pooling with predictions from other deep models. Ma et al. \cite{ma2017end} developed a network with two stages: the first stage determines the type of distortion, and the second stage predicts the subjective score using a quality prediction network specifically designed for each distortion type. Ahmed et al. \cite{ahmed2022deep} designed a CNN architecture by combining two architectures, resulting in a deeper and more complex model that demonstrated superior performance. They performed pre-training using pseudo-labeled images with simulated distortions and fine-tuned the model on a self-collected dataset of 12,000 authentically distorted images.\\
In a similar manner, some studies relied on the use of deep features for image quality evaluation and has shown promising results \cite{zhang2018unreasonable,tian2020light}. Regression methods trained on features extracted from pre-trained models like VGG or Residual networks have been developed to evaluate image quality \cite{ma2020multimedia}. Ahmed et al. \cite{ahmed2020perceptual} presented a study to explore the usefulness of deep features by training a regression algorithm for quality assessment. They demonstrated that fine-tuning significantly improves the predictive performance of deep features. Their final model consisted on Gaussian process regression, trained on NASNet-large deep features. In another study, Ahmed et al. \cite{ahmed2020image} used deep features along with natural scene statistics to construct a hybrid feature set for training quality regression. They demonstrated that augmenting deep features with suitably selected natural scene statistics can results in improved performance.\\
In a recent study, Aslam et al. \cite{aslam2023vrl} introduced a visual representation learning approach for IQA, performing weakly supervised pre-training on a dataset with synthetic distortions and fine-tuning on benchmark IQA datasets to learn specifics of authentic distortions. In \cite{aslam2024tqp}, they have used EfficientNet-B0 and used it for video quality assessment by introducing temporal shift to capture temporal element.\\
Ma et al. \cite{ma2021blind} proposed evaluating image quality using active inference, where the human visual system extrapolates relevant information from images for understanding. Their method employs active inference through main content prediction using a generative adversarial network, followed by image quality evaluation based on multi-stream CNNs trained on correlations between original and distorted images.
\subsection*{Integration of Various Models}
Ensemble learning, a popular approach to combine various learners for superior predictive performance, has also been explored extensively in the context of IQA. Ahmed et al. \cite{ahmed2019ensembling} proposed an ensemble learning method based on model snapshots obtained through a cyclic learning rate scheduler. This method intelligently combines model snapshots to construct an ensemble that outperforms individual models. In another study \cite{ahmed2022deep}, they have used Inception-ResNet-V2 and  EfficientNet-B7 to learn quality aware features which are concatenated and provided to multi-layer perceptron for image quality regression.\\
Apart from these methods, several other studies have also investigated ensemble learning for IQA. Zhang et al. \cite{zhang2021learning} developed a stacking ensemble of deep neural networks for no-reference IQA by training multiple CNNs on different subsets of distorted images and then combining their predictions using a weighted average method. This approach improved the robustness and accuracy of the quality predictions across various types of distortions.\\
In recent study \cite{wei2022perceptual}, an enhanced meta-learning framework is introduced to improve no-reference IQA by pre-training a meta-model and optimizer on diverse tasks with various distortions. This approach enabled the model to learn effective weight initialization and optimization rules, enhancing its ability to generalize and adapt to new tasks with minimal fine-tuning. Similarly, Zhu et al. \cite{zhu2020metaiqa} introduced MetaIQA, aiming to overcome the limitations of traditional CNN-based methods due to the small sample problem in IQA. By leveraging meta-learning, the model learns meta-knowledge from diverse distortion types, enabling rapid adaptation to new and unknown distortions. \\
Wang et al. \cite{wang2021semi} proposed a hierarchical ensemble learning framework that integrates multiple CNN models at different levels of abstraction. This framework first trains individual CNNs on different image quality aspects, such as sharpness, contrast, and color fidelity. The predictions from these CNNs are then combined using a higher-level ensemble model to produce a final image quality score. This hierarchical approach leverages the strengths of each individual CNN, resulting in more accurate and reliable quality assessments.\\
These studies demonstrate the effectiveness of ensemble learning and meta-learning techniques in improving IQA. By combining predictions from multiple models and leveraging meta-learning strategies, these approaches address the limitations of single-model methods, such as overfitting and lack of generalization. The advancements in ensemble learning and meta-learning highlight the potential for developing more accurate and reliable IQA systems that can better handle the diverse and complex nature of real-world image distortions.
\subsection*{Research Gap}
Despite advances, no-reference IQA faces challenges, particularly in generalization due to models focusing on specific quality aspects. Most approaches lack quality-aware pre-training, optimized loss functions for quality aware training, and have underexplored strategies to optimally combine multiple learners to improve generalization and predictive performance. This leaves a critical research gap in developing robust, adaptable models for authentically distorted images.\\
This study addresses these gaps by integrating quality-aware pre-training, novel loss function, meta-learning to leverage the power of multiple base learners modeled for authentically distorted image datasets such as Live in the Wild \cite{ghadiyaram2015massive}, KonIQ-10k \cite{hosu2020koniq}, and BIQ2021 \cite{ahmed2022biq2021}. These contributions aim to enhance IQA model accuracy and generalizability for real-world applications.
\section{Proposed Approach}
This section presents the proposed MetaQAP framework, a no-reference Image Quality Assessment (IQA) method that integrates quality-aware pretraining, meta-learning, and cross-dataset fine-tuning to enhance generalization on authentically distorted images. The approach is designed to systematically bridge the gap between synthetic pretraining and real-world evaluation through an iterative, data-driven learning process.
\subsection*{Research Framework}
To ensure a structured and reproducible workflow, this study adapts the CRISP-DM (Cross-Industry Standard Process for Data Mining) framework~\cite{wirth2000crisp}. While other methodologies such as KDD (Knowledge Discovery in Databases) and SEMMA (Sample, Explore, Modify, Model, Assess) provide systematic approaches to data-driven modeling, CRISP-DM offers greater flexibility and supports iterative refinement—an essential feature for tasks like IQA, where continuous feedback and re-evaluation are required. 
The proposed adaptation of CRISP-DM for IQA comprises five iterative stages: (1) data understanding, (2) data preparation, (3) modeling, (4) evaluation, and (5) deployment. This structure enables ongoing optimization of feature representations and model generalization.
\subsection*{Problem Understanding}
Image Quality Assessment (IQA) aims to predict human-perceived visual quality without access to a reference image. Existing no-reference (NR) methods often struggle to generalize across datasets, particularly when faced with authentic distortions that differ from synthetic training samples. 
To address these issues, this study introduces MetaQAP, a NR-IQA framework that combines (i) a quality-aware pretraining stage using synthetically distorted data, (ii) a fine-tuning stage on authentically distorted datasets, and (iii) a meta-learning phase that enhances cross-dataset generalization. The model is trained to align predicted quality scores with human Mean Opinion Scores (MOS), and its performance is measured using Pearson Linear Correlation Coefficient (PLCC) and Spearman Rank Order Correlation Coefficient (SROCC).
\subsection*{Dataset Understanding} \label{sec:dataset}
The proposed approach is evaluated using three benchmark datasets containing authentically distorted images: LIVE Challenge (LiveCD)~\cite{ghadiyaram2015massive}, KonIQ-10K~\cite{hosu2020koniq}, and BIQ2021~\cite{ahmed2022biq2021}. Each dataset includes Mean Opinion Scores (MOS) and variance values derived from large-scale human studies, providing reliable ground truth for perceptual quality assessment.
\subsubsection*{LiveCD}
\begin{figure}[htbp]
\centering
\includegraphics[width=0.5\linewidth]{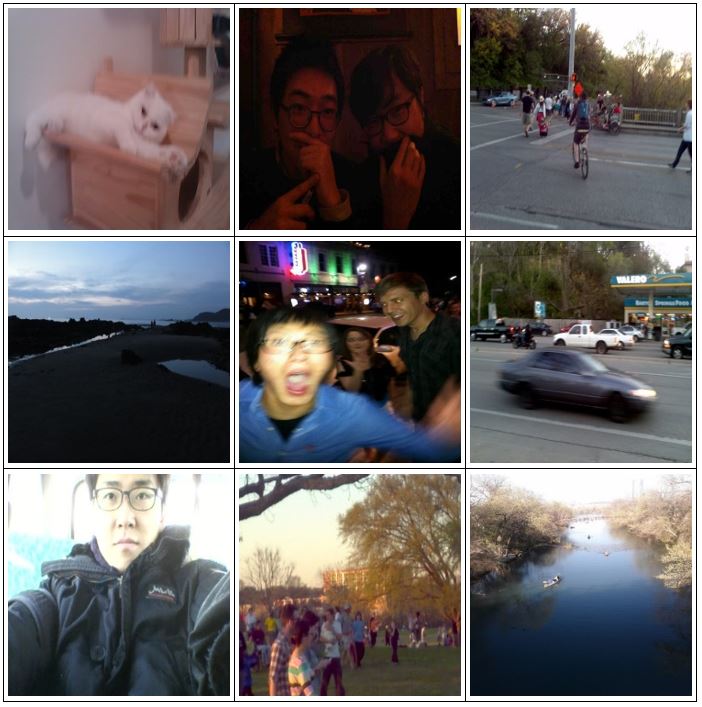}
\caption{Randomly selected images from the LiveCD dataset~\cite{ghadiyaram2015massive}}
\label{fig:LiveCD}
\end{figure}

The LiveCD dataset \cite{ghadiyaram2015massive} contains 1,162 images with authentic distortions captured by various mobile devices, reflecting real-world quality challenges. Over 8,100 participants on Amazon Mechanical Turk provided 350,000 quality ratings, with an average of 175 evaluations per image. The dataset includes both MOS (Mean Opinion Score) and variance to support robust quality assessments. Figure \ref{fig:LiveCD} shows randomly sampled images, illustrating its diversity in content and perceptual quality.
\subsubsection*{KonIQ-10K}
\begin{figure}[htbp]
\centering
\includegraphics[width=0.5\linewidth]{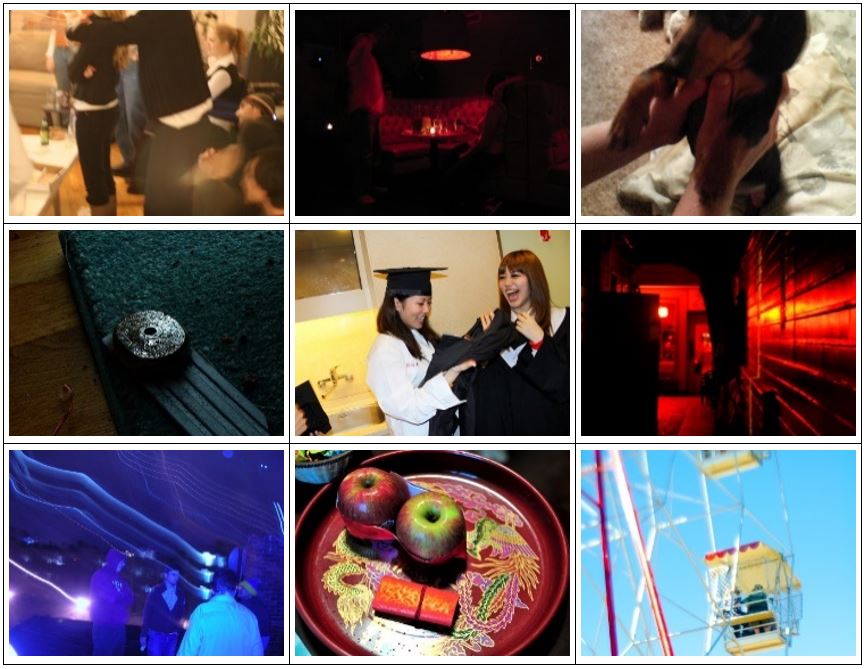}
\caption{Randomly sampled images from KonIQ-10K datasets~\cite{hosu2020koniq}}
\label{fig:KonIQ}
\end{figure}

KonIQ-10K \cite{hosu2020koniq} comprises 10,073 images selected from the YFCC100m database, designed to ensure ecological validity and diversity in content and quality. Quality ratings were collected from 1,467 individuals, resulting in 1.2 million evaluations. This dataset spans a wide spectrum of perceptual qualities and content types, as depicted in Figure \ref{fig:KonIQ}.
\subsubsection*{BIQ2021}
\begin{figure}[htbp]
\centering
\includegraphics[width=0.5\linewidth]{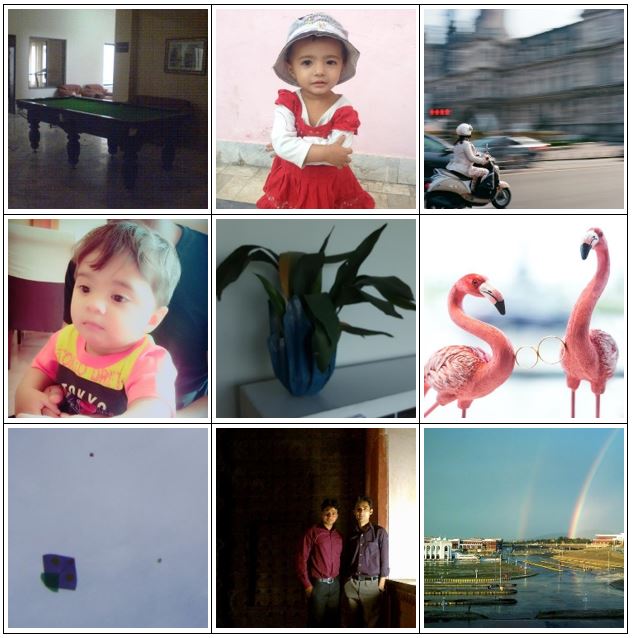}
\caption{Randomly selected images from the BIQ2021 dataset~\cite{ahmed2022biq2021}}
\label{fig:BIQ2021}
\end{figure}

BIQ2021 \cite{ahmed2022biq2021} includes 12,000 images with authentic distortions, categorized into three subsets: images captured under controlled conditions, by non-professional photographers, and by professionals. Each image was rated by 30 unique observers, totaling 360,000 evaluations. MOS and variance are provided for each image, enabling comprehensive training and evaluation. Figure \ref{fig:BIQ2021} highlights the dataset’s diversity in content and quality.
\subsection*{Data Preparation}
Data preparation is crucial for ensuring the robustness and generalizability of deep learning-based IQA models. Unlike traditional natural scene statistics approaches, CNNs automatically learn hierarchical representations of image quality from data. Following prior research~\cite{ravela2019no,ma2017end,ahmed2019ensembling,ahmed2022deep,zhang2020blind}, this study has strategically implemented preprocessing techniques to optimize model convergence and stability.
\subsubsection*{Image Resizing, Normalization, and Augmentation}
To ensure consistency in feature extraction, images are resized to a fixed dimension while maintaining aspect ratio. Instead of global resizing—which may alter perceptual quality—the model uses random cropping to extract multiple patches from each image. This technique acts as data augmentation by exposing the model to diverse regions of the same image, thereby improving generalization without altering perceptual integrity.
Each image is a 24-bit RGB input with pixel values normalized to the [0,1] range by dividing by 255, which stabilizes gradient updates and accelerates training. Additional augmentations, including translation and rotation, further expand the dataset without compromising perceived quality (Figure~\ref{fig:augmentation}).
\subsubsection*{Rescaling MOS}
The Mean Opinion Score (MOS) varies across datasets due to differing rating scales. To ensure consistent training and evaluation, MOS values are normalized using min–max scaling:
\begin{equation} \label{eq:mos}
    MOS_{scaled} = \frac{MOS - MOS_{min}}{MOS_{max} - MOS_{min}}
\end{equation}
This transformation standardizes all MOS values to the [0,1] range, where 0 represents the lowest perceptual quality and 1 the highest. Normalization facilitates smooth convergence during training and enables seamless integration of multiple datasets.
\subsubsection*{Data Augmentation}
Deep learning models rely on large datasets to achieve optimal performance, but acquiring extensive IQA-specific data is challenging. Image augmentation addresses this limitation by synthetically expanding the dataset through controlled variations. These augmentations not only increase the dataset size but also enhance the model’s ability to generalize by exposing it to diverse image scenarios, reducing the risk of overfitting.\\
In IQA, traditional augmentation techniques like scaling or contrast adjustments commonly used in classification tasks are unsuitable because they may alter the perceptual quality of images. Instead, this study employs carefully selected augmentation methods, including translation, rotation, and random cropping, which preserve perceptual quality while introducing meaningful variability. For instance, random cropping ensures the model encounters different regions of the same image during training, promoting a focus on perceptual quality rather than content-specific details.
\begin{figure}[htbp]
\centering
\includegraphics[width=0.7\linewidth]{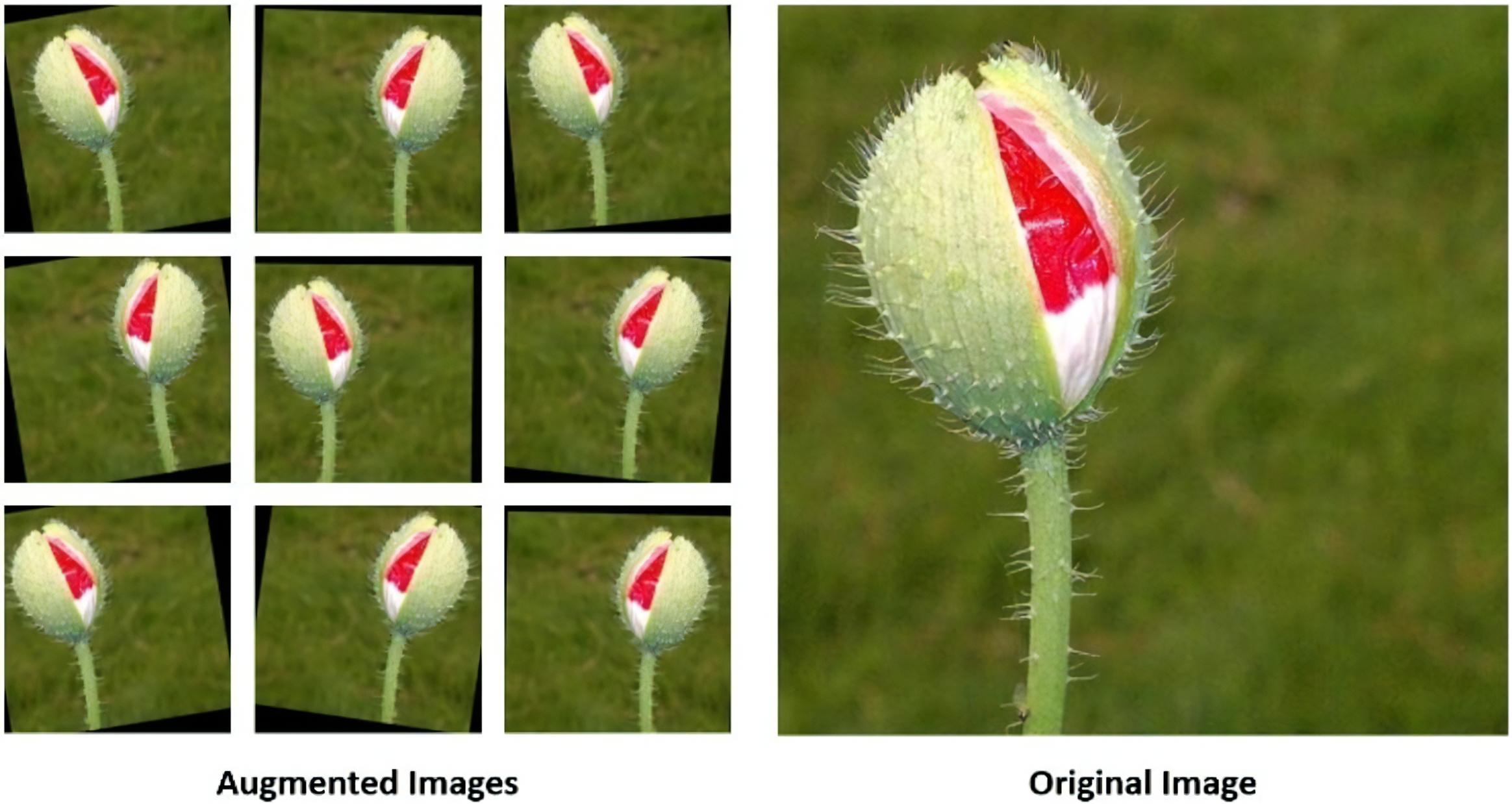}
\caption{Original image and its augmented variants generated by performing augmentation operations~\cite{ahmed2022biq2021}}
\label{fig:augmentation}
\end{figure}
The impact of these augmentations is illustrated in Figure \ref{fig:augmentation}, where the original image (right) and its augmented variants (left) demonstrate how the process enriches the training set. These techniques contribute to the robustness and reliability of the proposed model by simulating real-world variations without compromising the integrity of the perceptual quality assessment.
\subsubsection*{Data Partitioning}
To train and evaluate the model effectively, the dataset is divided into training and validation sets. Two validation strategies are commonly used: Cross-Validation is an iterative method divides the dataset into multiple folds, alternating training and validation splits. Although cross-validation is comprehensive, it is computationally expensive. Conversely, holdout Validation involves splitting the dataset, typically with a larger portion for training and a smaller portion for validation.\\
For model training, 80\% of each dataset is used for training and 20\% for validation. Although cross-validation offers more exhaustive evaluation, it is computationally demanding; therefore, the holdout method is adopted for efficiency. 
In cross-dataset evaluation, the model is trained on one dataset (e.g. KonIQ-10K) and tested on a distinct dataset (e.g., LIVE Challenge or BIQ2021) to rigorously assess generalization across unseen domains.
\subsection*{Model Building}
Model building is a crucial step in developing a robust IQA framework. Previous approaches, such as fine-tuning pre-trained models or using ensemble techniques, often fall short when applied to real-world scenarios, as they rely on synthetic datasets like LIVE \cite{sheikh2005live} and TID2013 \cite{ponomarenko2015image}that fail to capture the complexities of authentic distortions. This study addresses these limitations by focusing on datasets with authentically distorted images, reflecting real-world challenges such as acquisition flaws and post-processing artifacts.\\
A significant contribution of this work is the introduction of quality-aware pre-training, which equips the model with features directly relevant to perceptual quality. Additionally, a novel quality-aware loss function is designed to align predictions with human judgment by reducing errors while preserving perceptual ranking. Transfer learning accelerates convergence and adapts the model efficiently to IQA tasks with limited training data, while an integrated meta-learning framework ensures generalizability across diverse datasets and scenarios. These contributions collectively establish a framework designed to achieve accurate, robust and perceptually aligned IQA predictions, as demonstrated by the empirical evaluations presented in Section \ref{sec:results_discussion}, where our approach is validated through extensive experiments and comparative analysis.
\subsubsection*{Pre-Training on Quality-aware Dataset}
\begin{figure}[htbp]
\centering
\includegraphics[width=0.8\linewidth]{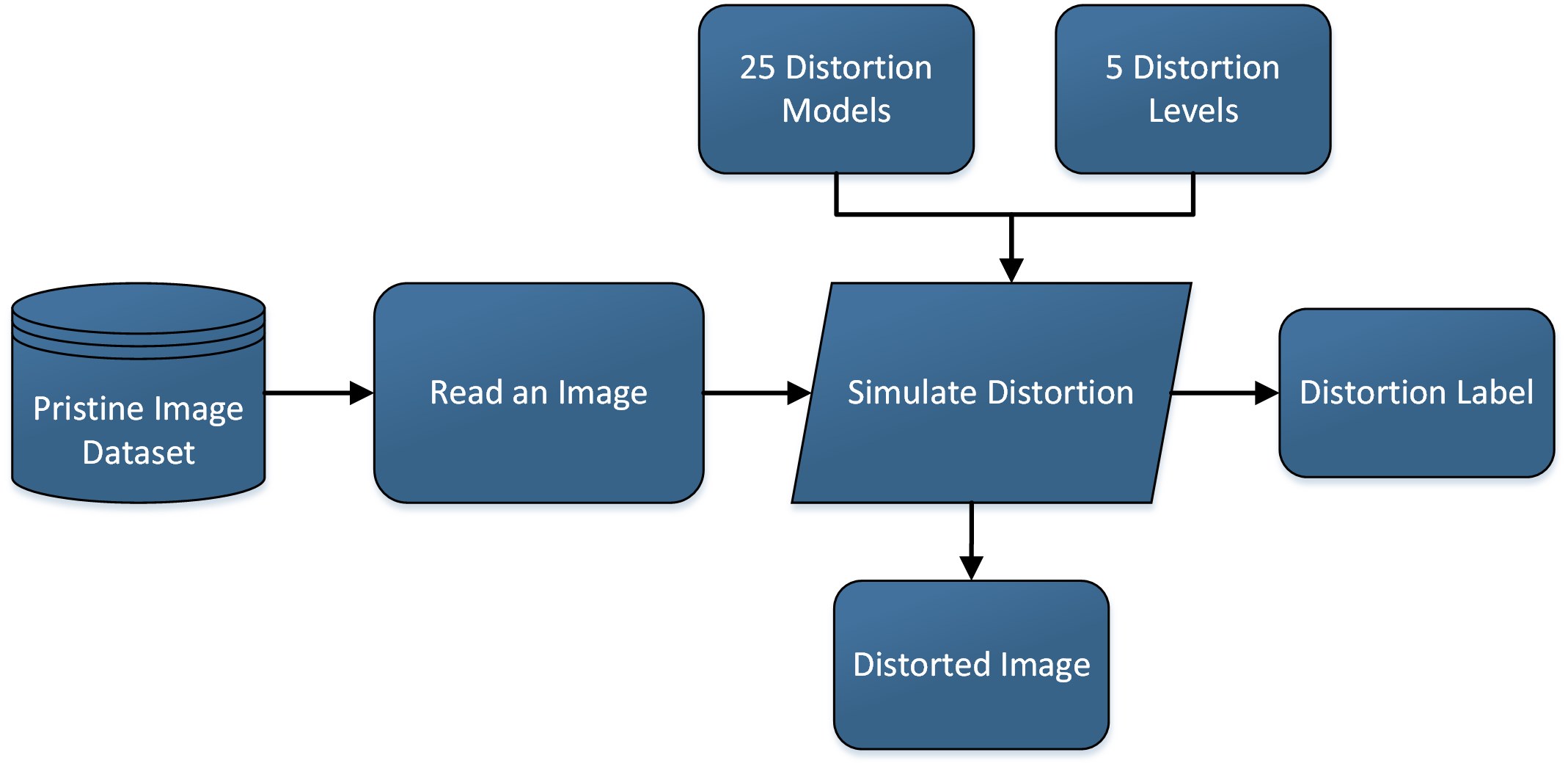}
\caption{Block diagram of distorted image generation and quality aware pre-training}
\label{fig:pre-training}
\end{figure}

Training CNNs for IQA is particularly challenging due to the limited availability of annotated datasets. While transfer learning, typically involving the fine-tuning of ImageNet-pretrained models, has been widely used to address this limitation, it often fails to fully capture the nuances of perceptual quality. This study introduces a novel pre-training approach, leveraging a custom quality-aware dataset to significantly enhance the model's ability to evaluate image quality with higher accuracy.\\
To create this dataset, we curated 200,000 pristine images from diverse sources, including Kadis-700K \cite{lin2019kadid}, the Waterloo Exploration Dataset \cite{ma2016waterloo}, and Unsplash.com. By applying 25 distinct distortion models, each with five levels of degradation granularity, we generated a massive set of 25 million distorted images. These distortion models, ranging from Gaussian blur to JPEG compression and motion blur, were specifically selected to simulate a wide variety of real-world imperfections. The specific details of the distortion models are provided in Table \ref{tab:distortion_model}. Each image in this dataset was meticulously annotated based on its distortion type and severity, enabling precise training of the model. For instance, an image distorted by the 25th model at the highest degradation level was labeled as “25-5,” ensuring a structured and meaningful annotation system.\\
This quality-aware dataset formed the basis for training a distortion classification model using a categorical cross-entropy loss function. Figure \ref{fig:pre-training} illustrates the pipeline for generating distortions and pre-training the model. Through this process, the CNN learned to recognize and differentiate between various distortions and their severities, effectively embedding perceptual quality indicators into its feature representations.
Unlike conventional pre-training approaches that rely on unrelated datasets, our strategy equips the model with a nuanced understanding of image distortions directly relevant to IQA tasks. This pre-training stage establishes a robust foundation, allowing the model to achieve superior performance when fine-tuned on specific IQA datasets containing authentic distortions. By addressing the inherent challenges of limited annotated data and the complexity of real-world distortions, this contribution sets a new benchmark for pre-training strategies in the IQA domain.
\begin{table*}[!ht]
\centering
\caption{Distortion models used to simulate distortion in the images}
\begin{adjustbox}{width=\textwidth}
\begin{tabular}{lll}
\toprule
\textbf{Sr.} & \textbf{Distortion Model} & \textbf{Description}                                                                                                                    \\ 
\midrule
1            & Color quantization                             & Convert to indexed image using minimum variance quantization and dithering.                                                \\ 
2            & JPEG2000                                       & Applies standard JPEG2000 compression.                                                                                                                        \\ 
3            & Impulse noise                                  & Adds salt and pepper noise to the RGB image.                                                                                                                  \\ 
4            & Color block                                    & Inserts homogeneous random colored blocks at random locations in the image.                                                                                   \\ 
5            & Color shift                                    & Blend green channel randomly into the normalized gradient magnitude of the image. \\ 
6            & Color diffusion                                & Gaussian blurs the color channels (a and b) in the Lab color space.                                                                                           \\ 
7            & Multiplicative noise                           & Introduces speckle noise to the RGB image.                                                                                                                    \\ 
8            & Denoise                                        & Adds Gaussian noise and perform denoising using DuCNN.                                                     \\ 
9            & Gaussian blur                                  & Applies a filter with a variable Gaussian kernel to blur the image.                                                                                           \\ 
10           & High sharpen                                   & Over-sharpens the image using unsharp masking.                                                                                                                \\ 
11           & Darken                                         & Similar to brighten but decreases other luminance values to darken the image.                                                                                 \\ 
12           & Contrast change                                & Non-linearly change the color contrast of the image in the form of S-curve.                                                                                   \\ 
13           & HSV Color saturation                           & Multiplies the saturation channel in the HSV color space by a factor.                                                                                         \\ 
14           & Jitter                                         & Randomly scatter image data with random small offsets using bicubic interpolation.                                                     \\ 
15           & Lens blur                                      & Uses a filter with a circular kernel to simulate lens blur.                                                                                                   \\ 
16           & JPEG                                           & Uses standard JPEG compression.                                                                                                                               \\ 
17           & Mean shift                                     & Adds a constant value to all pixels and truncates to the original value range.                                                                                \\ 
18           & Brighten                                       & Non-linearly adjusts the luminance channel to increase brightness.                                                         \\
19           & Lab Color saturation                           & Multiplies the color channels in the Lab color space by a factor.                                                                                             \\ 
20           & Non-eccentricity patch                         & Randomly offsets small patches in the image to nearby locations.                                                                                              \\ 
21           & Pixelate                                       & Downscale the image \& perform upsampling using nearest-neighbor interpolation.                                        \\ 
22           & Quantization                                   & Quantizes image values using N thresholds obtained with Otsu’s method.                                                                                        \\ 
23           & Motion blur                                    & Applies a filter with a line kernel to create motion blur effects.                                                                                            \\ 
24           & White noise in YCbCr                           & Adds Gaussian noise to the luminance \& chrominance channels of the YCbCr image.                                                   \\ 
25           & White noise                                    & Adds Gaussian white noise to the RGB image.                                                                                                                   \\ 
\bottomrule
\end{tabular}
\label{tab:distortion_model}
\end{adjustbox}
\end{table*}
\subsubsection*{Quality-Aware Loss Function}
The choice of an appropriate loss function is pivotal in the training of IQA models, as it directly influences their ability to predict perceptual quality accurately. While MSE remains the most commonly used loss function due to its simplicity and focus on minimizing prediction errors, it fails to capture the nuanced relationship between predicted and ground truth scores \cite{hosu2020koniq,ahmed2022deep,aslam2024tqp}. Recognizing this limitation, this study introduces a novel quality-aware loss function that integrates both error-based and correlation-based metrics to enhance the predictive performance of IQA models.\\
The core innovation lies in combining MSE with a differentiable approximation of the SROCC. This dual-component loss function addresses two key objectives: minimizing the magnitude of prediction errors and maintaining the monotonic relationship between predicted and ground truth quality scores. While MSE focuses on reducing discrepancies in absolute values, the SROCC component ensures the relative rankings of predictions align with human perceptual judgments. This combination provides a balanced approach, capturing both the quantitative and perceptual aspects of image quality.\\
The proposed quality-aware loss function $QA_{loss}$ integrates MSE and a differentiable approximation of SROCC. The $QA_{loss}$ is defined as (\ref{eq:loss}):
\begin{equation} \label{eq:loss}
    QA_{loss} = \lambda_1 \cdot MSE +  \lambda_2 \cdot (1 - SROCC)
\end{equation}
In this formulation, MSE (defined by \ref{eq:mse}) quantifies the average squared difference between predicted and ground truth quality scores:
\begin{equation} \label{eq:mse}
    \text{MSE} = \frac{1}{2} \cdot \sum_{i=1}^{n} (Y_{i}-\hat{Y_{i}})^2
\end{equation}
where $Y_{i}$ and $\hat{Y_{i}}$ denote the ground truth and predicted scores, respectively.\\
To incorporate SROCC into the loss function, we address its inherent non-differentiability by leveraging a softmax-based approximation (\ref{eq:srocc_aprox}), enabling smooth optimization during training.
\begin{equation} \label{eq:srocc_aprox}
    R_{i} = \frac{e^{Y_i}}{\sum_{j=1}^{n} e^{Y_j}}
\end{equation}
The differentiable SROCC component is expressed as $1-SROCC$ to frame it as a loss term (i.e., minimizing it is desirable). The coefficients $\lambda_1$ and $\lambda_2$ control the relative importance of the MSE and SROCC components, respectively.\\
This novel formulation allows the model to learn both to reduce prediction errors and to preserve the relative quality rankings of images, offering a more perceptually aligned training objective.
\subsubsection*{Optimization of Loss Hyperparameters}
To maximize the effectiveness of the proposed quality-aware loss function, the weighting coefficients 
$\lambda_1$ and $\lambda_2$ were systematically optimized using a grid search approach. A total of 16 combinations of 
$\lambda_1$ and $\lambda_2$ values were evaluated, with each configuration tested over multiple training epochs. The performance of the loss function for each combination was assessed, and the results are summarized in Table~\ref{tab:loss}.
\begin{table}[ht!]
\centering
\caption{Results of loss values at 16 grid points}
\begin{tabular}{ccccc}
\toprule
\textbf{Parameters} & \textbf{$\lambda_1=\frac{1}{4}$} & \textbf{$\lambda_1=\frac{1}{2}$} & \textbf{$\lambda_1=\frac{3}{4}$} & \textbf{$\lambda_1=1$} \\ 
\midrule
\textbf{$\lambda_2=\frac{1}{4}$}         & 0.042        & 0.036        & 0.037        & 0.043      \\ 
\textbf{$\lambda_2=\frac{1}{2}$}         & 0.038        & \textbf{0.032}        & 0.035        & 0.039      \\ 
\textbf{$\lambda_2=\frac{3}{4}$}         & 0.041        & 0.034        & 0.033        & 0.037      \\ 
\textbf{$\lambda_2=1$}           & 0.045        & 0.039        & 0.04         & 0.041      \\ 
\bottomrule
\end{tabular}
\label{tab:loss}
\end{table}
The optimal configuration, identified as $\lambda_1 = \frac{1}{2}$ and $\lambda_2 = \frac{1}{2}$, yielded the lowest loss value of 0.032. This balance between error reduction and correlation maximization proved most effective for enhancing the model’s generalization and predictive accuracy. By assigning equal importance to both components, the quality-aware loss function achieves a nuanced optimization objective, capturing both the magnitude and perceptual relationships inherent in image quality assessment.\\
This contribution perform integration of error- and correlation-based metrics offers a comprehensive solution that not only improves prediction accuracy but also ensures alignment with human perceptual judgments. The proposed $\text{QA}_{\text{loss}}$ function provides a robust and effective training objective, paving the way for the development of highly generalizable and perceptually accurate IQA models.
\subsection*{Fine-Tuning}
\begin{figure}[htbp]
\centering
\includegraphics[width=0.6\linewidth]{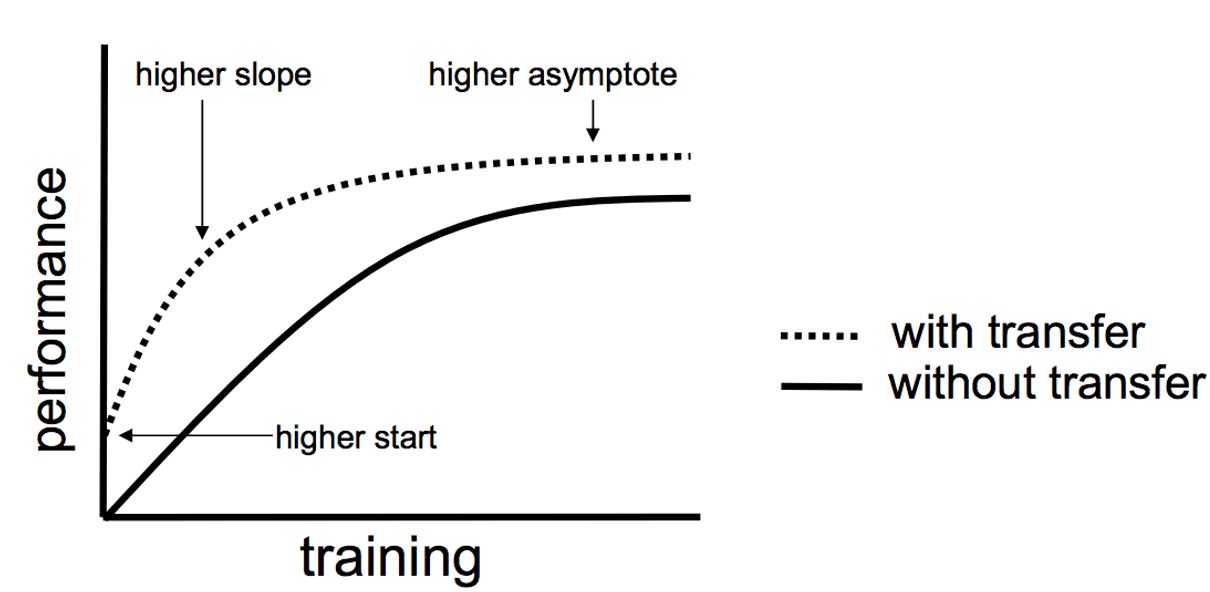}
\caption{Model training performance with and without transfer learning}
\label{fig:transfer}
\end{figure}

Transfer learning aims to leverage a model pre-trained on one task and fine-tune it for a similar, yet distinct, task. This approach, rooted in the principles of knowledge transfer from psychological literature, significantly enhances sample efficiency in prediction problems. By leveraging models pre-trained on large datasets, transfer learning significantly reduces the need for extensive labeled data and enhances the efficiency of training. In the context of IQA, this approach allows a classification model pre-trained on a large-scale distortion dataset to serve as a strong foundation for fine-tuning on authentically distorted datasets. These pre-trained models carry forward quality-aware features, enabling the effective learning of nuanced distortion patterns even with smaller training datasets.\\
The effectiveness of transfer learning in IQA is underscored by its ability to deliver three critical advantages: higher initial performance compared to training from scratch, steeper learning curves during training, and superior convergence performance. As shown in Figure \ref{fig:transfer}, the training process with transfer learning not only accelerates model optimization but also achieves higher predictive accuracy, validating its efficacy for IQA tasks.
\subsubsection*{Proposed Model} \label{sec:base_learners}
To identify the most suitable architecture for IQA, various state-of-the-art CNN models were fine-tuned on authentically distorted datasets. The classification layers of these models were restructured to accommodate regression tasks by replacing softmax layers with a regression layer comprising a single neuron. This adaptation, coupled with the use of the quality-aware loss function, ensured that the models could predict image quality scores with high alignment to human perception. Table \ref{tab:base_learners} presents the performance of individual CNN architectures on three benchmark datasets, measured using PLCC and SROCC metrics.\\
The results reveal that models such as NasNet\_Large and DenseNet-201 outperform others in terms of correlation with human judgments, showcasing their ability to generalize well across authentically distorted datasets. However, these models also entail significantly higher computational complexity compared to others. This exploration of fine-tuning techniques not only highlights the adaptability of pre-trained models but also demonstrates how leveraging pre-existing quality-aware features can significantly elevate the performance of IQA systems. By refining network architectures and employing a comprehensive loss function, this study sets a benchmark for using transfer learning in achieving robust and perceptually aligned IQA models.
\begin{table*}[htbp]
\centering
\caption{Quality Prediction Performance of Individual CNN Models}
\label{tab:base_learners}

\begin{adjustbox}{width=\textwidth}
\begin{threeparttable}
\begin{tabular}{
c l c c c c
cc cc cc
}
\toprule
\multirow{2}{*}{\textbf{Sr.}} &
\multirow{2}{*}{\textbf{Network}} &
\textbf{Parameters} &
\textbf{Memory} &
\textbf{Depth} &
\textbf{MACs} &
\multicolumn{2}{c}{\textbf{KonIQ-10K}} &
\multicolumn{2}{c}{\textbf{LiveCD}} &
\multicolumn{2}{c}{\textbf{BIQ2021}} \\
\cmidrule(lr){7-12}
 & & \textbf{(Millions)} & \textbf{(MB)} & \textbf{(Layers)} & \textbf{(GMACs)} &
 \textbf{PLCC} & \textbf{SROCC} &
 \textbf{PLCC} & \textbf{SROCC} &
 \textbf{PLCC} & \textbf{SROCC} \\
\midrule
1 & Densenet-201 & 20.0 & 77 & 201 & $\sim$4.4 & 0.9404 & 0.9268 & 0.9093 & 0.8957 & 0.7465 & 0.7329 \\
2 & Efficientnet-b0 & 5.3 & 20 & 82 & $\sim$0.39 & 0.8509 & 0.8265 & 0.8096 & 0.7852 & 0.5995 & 0.5773 \\ 
3 & GoogleNet & 7.0 & 27 & 22 & $\sim$1.5 & 0.8867 & 0.8791 & 0.8165 & 0.8089 & 0.6249 & 0.6005 \\ 
4 & Inception-V3 & 23.9 & 91 & 48 & $\sim$5.7 & 0.5206 & 0.6874 & 0.5528 & 0.7196 & 0.6808 & 0.6732 \\ 
5 & InceptionResNet-V2 & 55.9 & 213 & 164 & $\sim$13.2 & 0.6622 & 0.7833 & 0.6796 & 0.8007 & 0.3379 & 0.5047 \\
6 & MobileNet-V2 & 3.5 & 14 & 53 & $\sim$0.3 & 0.8626 & 0.8445 & 0.8318 & 0.8137 & 0.4989 & 0.6200 \\
7 & NasNet\_Large & 88.9 & 344 & - & $\sim$23.8 & 0.9643 & 0.9403 & 0.9717 & 0.9477 & 0.6843 & 0.6662 \\ 
8 & NasNet\_Mobile & 5.3 & 20 & - & $\sim$1.1 & 0.8195 & 0.8329 & 0.7477 & 0.7611 & 0.8505 & 0.8265 \\ 
9 & Resnet18 & 11.7 & 45 & 18 & $\sim$1.8 & 0.7855 & 0.7896 & 0.7563 & 0.7604 & 0.6067 & 0.6201 \\ 
10 & Resnet50 & 25.6 & 98 & 50 & $\sim$3.9 & 0.8328 & 0.8123 & 0.8483 & 0.8278 & 0.6091 & 0.6132 \\ 
11 & Resnet101 & 44.6 & 171 & 101 & $\sim$7.6 & 0.7119 & 0.7114 & 0.7271 & 0.7266 & 0.7267 & 0.7062 \\ 
12 & ShuffleNet & 1.4 & 5.5 & 50 & $\sim$0.15 & 0.7889 & 0.7843 & 0.7380 & 0.7334 & 0.5836 & 0.5831 \\ 
13 & SqueezeNet & 1.2 & 4.7 & 18 & $\sim$0.8 & 0.7664 & 0.7583 & 0.7703 & 0.7622 & 0.5481 & 0.5435 \\ 
14 & Xception & 22.9 & 88 & 71 & $\sim$8.4 & 0.9279 & 0.8997 & 0.9153 & 0.8871 & 0.5874 & 0.5793 \\

\bottomrule
\end{tabular}
\end{threeparttable}
\end{adjustbox}
\end{table*}

\subsubsection*{Meta Learning}
Meta-learning introduces an additional layer of intelligence to the ensemble of base models, optimizing their collective predictions for superior IQA performance. Let $M = \{M_1, M_2, \dots, M_n\}$ represent the candidate models, each modified with a global average pooling layer and a regression layer to generate quality predictions $P_i(X)$ for an input image $X$. These predictions are then aggregated using a Stepwise Linear Regression (SLR) model, which functions as the meta-learner, selectively combining predictions from the base models to produce the final quality prediction.\\
The SLR algorithm iteratively identifies the most relevant predictors, discarding models with negligible contributions, to construct a regression equation optimized for performance. To ensure computational efficiency and robust predictions, a threshold of $\beta \geq 0.05$ is applied, retaining only base learners with substantial regression coefficients. This threshold minimizes the inclusion of redundant or less impactful models, streamlining the meta-learner's operation without compromising accuracy. The resulting linear regression equation is expressed as (\ref{eq:regression}):
\begin{equation} \label{eq:regression}
Y = \beta_0 + \beta_1 P_1(X) + \beta_2 P_2(X) + \dots + \beta_n P_n(X) + \epsilon
\end{equation}
where $Y$ is the final quality prediction, $P_i(X)$ are the predictions of individual models, $\beta_i$ are the learned regression coefficients, and $\epsilon$ is the error term. The pseudo-code for the SLR-based meta-learning process is presented in Algorithm~\ref{alg:stepwise_regression}.\\
The stepwise regression algorithm ensures the optimal combination of predictors, balancing performance with computational efficiency. By iteratively fitting linear models and evaluating their coefficients and scores, the algorithm dynamically refines the meta-learner, selecting predictors that significantly improve accuracy. This process culminates in a compact and effective ensemble model, as demonstrated in the final regression equation (\ref{eq:final_regression}):
\begin{equation} \label{eq:final_regression}
Y = 0.05 + 0.40 \cdot P_7(X) + 0.35 \cdot P_1(X) + 0.15 \cdot P_{14}(X) + 0.10 \cdot P_{10}(X)
\end{equation}
Here, $P_7(X)$, $P_1(X)$, $P_{14}(X)$, and $P_{10}(X)$ correspond to the serial numbers of base learners in Table~\ref{tab:base_learners}. The coefficients highlight the relative importance of each model, with higher values indicating stronger contributions to the final prediction.\\
By leveraging SLR for meta-learning, this study introduces a novel ensemble strategy that not only improves prediction accuracy but also ensures interpretability and efficiency. The selective integration of base models maximizes the collective potential of the ensemble, offering a sophisticated approach to IQA that balances robustness with computational practicality. This contribution sets a benchmark for applying meta-learning in quality assessment tasks, paving the way for further innovations in ensemble modeling.
\begin{algorithm}
\caption{Meta Learning using SLR}
\label{alg:stepwise_regression}

\begin{algorithmic}[1]
\Require Predictions: List of predictions from individual models (all data points).
\Require Target: Ground truth quality labels (all data points).
\Require Threshold: Minimum acceptable absolute value for regression coefficient (0.05).

\Ensure Selected\_predictions: List of selected predictors for the final model.

\State Initialize: Selected\_predictions $\gets$ empty\_list
\State Initialize: Full\_model $\gets$ LinearRegression(Predictions, Target)
\State Full\_model.fit(Predictions, Target)

\While{True}
    \State Best\_predictor $\gets$ None
    \State Best\_score $\gets$ $-\infty$  

    \ForAll{prediction $\in$ Predictions \textbackslash Selected\_predictions}
        \State New\_model $\gets$ LinearRegression()
        \State New\_model.fit(Predictions[\{prediction\} $\cup$ Selected\_predictions], Target)
        \State Score $\gets$ New\_model.score(Predictions[\{prediction\} $\cup$ Selected\_predictions], Target)
        \If{$|\text{New\_model.coef}[0]| \geq$ Threshold \textbf{and} Score $>$ Best\_score}
            \State Best\_predictor $\gets$ prediction
            \State Best\_score $\gets$ Score
        \EndIf
    \EndFor

    \If{Best\_predictor is None} \textbf{break} \EndIf

    \State Append Best\_predictor to Selected\_predictions
    \State Update Full\_model: Full\_model.fit(Predictions[Selected\_predictions], Target)
\EndWhile

\State \textbf{return} Selected\_predictions
\end{algorithmic}
\end{algorithm}
\section{Experimental Setup}
The experimental setup establishes the foundation for rigorously evaluating and comparing the performance of IQA models. This section details the execution environment, training parameters, and evaluation methodologies employed to ensure unbiased and comprehensive assessments. By meticulously defining these elements, the study guarantees reproducibility and provides a robust framework for benchmarking IQA models across datasets and evaluation metrics.
\subsection*{Execution Environment}
The implementation and evaluation of models were conducted using MATLAB ® 2024a on a 64-bit Windows 10 platform. A Dell T5600 workstation powered the computational tasks, featuring dual Intel Xeon E5-2587 processors, 32 GB of RAM, and a 512 GB SSD. For GPU-accelerated computations, an NVIDIA GTX 1070 Ti with 8 GB of GDDR5 memory was employed. This configuration provided the necessary computational resources to train deep neural networks efficiently, ensuring a balance between performance and cost-effectiveness.
\subsection*{Training Parameters}
Careful selection and optimization of training parameters were pivotal to achieving optimal model performance. Hyperparameter tuning was employed to determine the best configuration for each CNN model, considering variations in model size and dataset characteristics. The training parameters, summarized in Table \ref{tab:hyperparameters}, included the optimizer, mini-batch size, learning rate, and validation criteria.
\begin{table}[!ht]
\centering
\caption{Training parameters model training}
\begin{tabular}{llc}
\toprule
\textbf{Sr.} & \textbf{Parameter}          & \textbf{Value}       \\ 
\midrule
1            & Optimizer                   & Adam                 \\ 
2            & Mini Batch Size             & 8-64                 \\ 
3            & Max Epochs                  & 100                  \\ 
4            & Initial Learning Rate       & 0.001                \\ 
5            & Learning Rate Scheduler     & Piecewise (20 period)\\ 
6            & Validation Frequency        & 188-1500 (parience 10)\\ 
7            & LearnRateDropFactor         & 20                     \\
8            & LearnRateDropPeriod         & 20                     \\
9            & ValidationPatience          & 20                     \\
\bottomrule
\end{tabular}
\label{tab:hyperparameters}
\end{table}
These parameters were systematically tuned to optimize the balance between convergence speed and model accuracy. The use of a piecewise learning rate scheduler, paired with early stopping based on validation loss, ensured efficient utilization of computational resources while preventing overfitting.
\subsection*{Model Evaluation}
The evaluation of IQA models focuses on their ability to predict quality scores that align with human perceptual judgments. Unlike standard regression problems, where metrics such as MSE and RMSE quantify absolute prediction errors, IQA emphasizes the correlation between predicted scores and ground truth values. To this end, two widely adopted metrics, PLCC and SROCC, are used to evaluate model performance.
\subsubsection*{PLCC}
PLCC measures the strength and direction of the linear relationship between predicted and observed quality scores. It ranges from $-1$ (perfect negative correlation) to $+1$ (perfect positive correlation), with $0$ indicating no linear correlation. This metric evaluates how accurately the model captures linear dependencies in perceptual quality predictions.
\subsubsection*{SROCC}
SROCC evaluates the monotonic relationship between predicted and observed scores, assessing how well the rank order of predictions matches human judgments. Unlike PLCC, SROCC does not assume linearity and is less sensitive to outliers, making it ideal for capturing non-linear perceptual relationships.

By combining PLCC and SROCC, the evaluation framework provides a comprehensive assessment of IQA models. PLCC captures the precision of linear predictions, while SROCC emphasizes the model’s ability to rank images consistently with human perception. Together, these metrics offer a nuanced understanding of the model’s performance in predicting perceptual image quality.
\section{Results and Discussion} \label{sec:results_discussion}
This section provides a detailed evaluation of the proposed IQA model through extensive experiments designed to assess its performance. The evaluation includes training and fine-tuning CNN architectures, testing model generalizability, and conducting ablation studies to highlight the importance of each component in the framework.
\subsection*{Experimental Evaluation}
Four primary experiments were conducted:
\begin{enumerate}
    \item \textbf{Individual Model Evaluation:} Individual CNN architectures were trained and evaluated after pre-training and fine-tuning using the proposed methodology. Results are presented in Table \ref{tab:base_learners} (section \ref{sec:base_learners}).
    \item \textbf{Performance on Benchmark Datasets:} The proposed model was evaluated on three authentically distorted datasets using holdout validation to assess predictive accuracy and correlation with human perception.
    \item \textbf{Cross-Dataset Evaluation:} The generalization capability of the model was tested by training on one dataset and evaluating on others, ensuring robustness across varied data distributions.
    \item \textbf{Ablation Study:} The role of individual model components was analyzed by systematically removing or modifying them to determine their contributions to overall performance.
\end{enumerate}
\subsection*{Performance Comparison with Existing Schemes}
\begin{figure}
    \centering
    \includegraphics[width=0.8\linewidth]{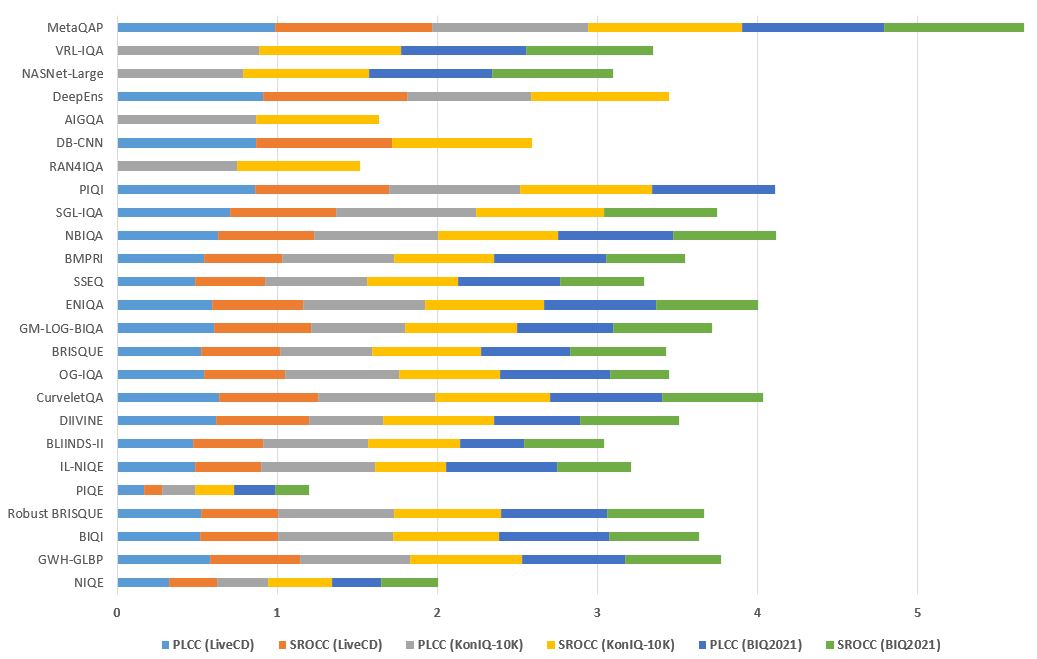}
    \caption{Comparison of various models using stacked bar-chart to highlight overall performance}
    \label{fig:stacked}
\end{figure}

The proposed model outperformed existing methods across all three benchmark datasets (LiveCD, KonIQ-10K, and BIQ2021) in both PLCC and SROCC metrics, as shown in Table \ref{tab:benchmark}. Compared to traditional no-reference IQA methods, such as NIQE, BIQI, and BRISQUE, the proposed model achieved substantial improvements, demonstrating its robustness in predicting quality under authentic distortions. For instance, the proposed model achieved a PLCC of 0.9885 on LiveCD, significantly surpassing the 0.8617 reported by PIQI, a close competitor. These results highlight the effectiveness of integrating pre-training, the quality-aware loss function, and the meta-learner in delivering state-of-the-art performance. These findings demonstrate the robustness and effectiveness of the proposed model across diverse benchmark datasets, highlighting its ability to accurately predict image quality scores under various authentic distortions. Figure \ref{fig:stacked} provide the stacked bar-chart depicting the performance of each method in terms of various metrics for an easier comparison.
\begin{table*}[!htbp]
\centering
\caption{Performance comparison with existing schemes}
\label{tab:benchmark}

\begin{adjustbox}{width=\textwidth}
\begin{threeparttable}
\begin{tabular}{
l
cc
cc
cc
}
\toprule
\textbf{Methods} &
\multicolumn{2}{c}{\textbf{LiveCD}} &
\multicolumn{2}{c}{\textbf{KonIQ-10K}} &
\multicolumn{2}{c}{\textbf{BIQ2021}} \\
\cmidrule(lr){2-7}
 & \textbf{PLCC} & \textbf{SROCC} &
   \textbf{PLCC} & \textbf{SROCC} &
   \textbf{PLCC} & \textbf{SROCC} \\
\midrule
NIQE \cite{mittal2012making}           & 0.328 & 0.299 & 0.319 & 0.400 & 0.301 & 0.356 \\
GWH-GLBP \cite{li2016no}               & 0.584 & 0.559 & 0.688 & 0.698 & 0.644 & 0.602 \\
BIQI \cite{moorthy2009modular}         & 0.519 & 0.488 & 0.718 & 0.662 & 0.683 & 0.564 \\
Robust BRISQUE \cite{mittal2012making} & 0.522 & 0.484 & 0.723 & 0.668 & 0.664 & 0.605 \\
PIQE \cite{venkatanath2015blind}       & 0.172 & 0.108 & 0.208 & 0.246 & 0.255 & 0.213 \\
IL-NIQE \cite{zhang2015feature}        & 0.487 & 0.415 & 0.707 & 0.447 & 0.694 & 0.461 \\
BLIINDS-II \cite{saad2012blind}        & 0.473 & 0.442 & 0.652 & 0.575 & 0.403 & 0.496 \\
DIIVINE \cite{moorthy2011blind}        & 0.617 & 0.580 & 0.463 & 0.693 & 0.541 & 0.617 \\
CurveletQA \cite{liu2014no}            & 0.636 & 0.621 & 0.730 & 0.718 & 0.698 & 0.630 \\
OG-IQA \cite{liu2016blind}             & 0.545 & 0.505 & 0.709 & 0.635 & 0.684 & 0.371 \\
BRISQUE \cite{mittal2012no}            & 0.524 & 0.497 & 0.574 & 0.677 & 0.555 & 0.603 \\
GM-LOG-BIQA \cite{xue2014blind}        & 0.607 & 0.604 & 0.589 & 0.696 & 0.603 & 0.617 \\
ENIQA \cite{chen2019no}                & 0.596 & 0.564 & 0.761 & 0.745 & 0.703 & 0.634 \\
SSEQ \cite{liu2014no}                  & 0.487 & 0.436 & 0.637 & 0.572 & 0.633 & 0.528 \\
BMPRI \cite{min2018blind}              & 0.541 & 0.487 & 0.705 & 0.619 & 0.699 & 0.494 \\
NBIQA \cite{ou2019novel}               & 0.629 & 0.604 & 0.771 & 0.749 & 0.718 & 0.642 \\
SGL-IQA \cite{varga2023no}             & 0.704 & 0.667 & 0.874 & 0.794 & -- & 0.710 \\
PIQI \cite{ahmed2021piqi}              & 0.862 & 0.838 & 0.815 & 0.824 & 0.728 & 0.745 \\
RAN4IQA \cite{ren2018ran4iqa}          & -- & -- & 0.752 & 0.763 & -- & -- \\
DB-CNN \cite{zhang2018blind}           & 0.869 & 0.851 & -- & 0.875 & -- & -- \\
AIGQA \cite{ma2021blind}               & -- & -- & 0.868 & 0.766 & -- & -- \\
DeepEns \cite{ahmed2022deep}           & 0.914 & 0.899 & 0.773 & 0.864 & 0.791 & 0.881 \\
NASNet-Large \cite{ahmed2022deep}      & 0.837 & 0.799 & 0.790 & 0.785 & 0.770 & 0.750 \\
VRL-IQA \cite{aslam2023vrl}            & 0.941 & 0.905 & 0.887 & 0.884 & 0.786 & 0.793 \\
IQDLNet \cite{xie2023no}               & 0.873 & 0.869 & 0.921 & 0.907 & -- & -- \\
TReS \cite{golestaneh2022no}           & 0.877 & 0.846 & 0.928 & 0.915 & -- & -- \\
ARNIQA \cite{agnolucci2024arniqa}      & 0.942 & 0.937 & 0.893 & 0.870 & 0.845 & 0.833 \\
UNI-IQA \cite{song2025uni}             & 0.885 & 0.874 & 0.937 & 0.928 & 0.898 & 0.907 \\
Q-Align \cite{wu2023q}                 & 0.887 & 0.883 & 0.935 & 0.934 & -- & -- \\
\textbf{MetaQAP (Ours)}                & \textbf{0.988} & \textbf{0.981} & \textbf{0.970} & \textbf{0.966} & \textbf{0.884} & \textbf{0.876} \\
\bottomrule
\end{tabular}
\end{threeparttable}
\end{adjustbox}
\end{table*}

\subsection*{Cross-Dataset Evaluation}
Cross-dataset evaluation is a crucial measure of a model’s generalizability, particularly in real-world applications where distortions vary significantly across datasets. Table \ref{tab:cross_dataset} presents the results of our cross-dataset experiments, where the proposed model was trained on one dataset and evaluated on another to ensure there is no overlap between training and testing data. The results demonstrate that models trained on larger and more diverse datasets, such as KonIQ-10K and BIQ2021, achieve higher PLCC and SROCC values on unseen datasets compared to those trained on smaller datasets like LiveCD. For instance, the model trained on BIQ2021 achieved a PLCC of 0.8023 on KonIQ-10K, reflecting its improved ability to generalize across different data distributions.
\begin{table}[!ht]
\centering
\caption{Model evaluation on cross-dataset scenario}
\begin{tabular}{llcc}
\toprule
\textbf{Training   Dataset} & \textbf{Testing Dataset} & \textbf{PLCC} & \textbf{SROCC} \\ 
\midrule
KonIQ-10K \cite{hosu2020koniq}                  & LiveCD \cite{ghadiyaram2015massive}                  & 0.7729        & 0.7489         \\ 
KonIQ-10K  \cite{hosu2020koniq}                 & BIQ2021 \cite{ahmed2022biq2021}                  & 0.7845        & 0.7588         \\ 
BIQ2021  \cite{ahmed2022biq2021}                   & LiveCD \cite{ghadiyaram2015massive}                  & 0.7910        & 0.7702         \\ 
BIQ2021  \cite{ahmed2022biq2021}                   & KonIQ-10K  \cite{hosu2020koniq}              & 0.8023        & 0.7805         \\ 
LiveCD \cite{ghadiyaram2015massive}                     & KonIQ-10K  \cite{hosu2020koniq}              & 0.6847        & 0.6629         \\ 
LiveCD  \cite{ghadiyaram2015massive}                    & BIQ2021 \cite{ahmed2022biq2021}                  & 0.6721        & 0.6515         \\ 
\bottomrule
\end{tabular}
\label{tab:cross_dataset}
\end{table}
The observed performance gap between in-dataset evaluations (Table \ref{tab:benchmark}) and cross-dataset evaluations (Table \ref{tab:cross_dataset}) highlights a fundamental challenge in no-reference IQA. Due to the intricacy and unpredictability of real distortions, models frequently learn dataset-specific characteristics, which limits their capacity to generalize to new data. This problem is specific to no-reference IQA models and emphasizes how crucial it is to create strong architectures that can reduce dataset dependency. The suggested ensemble approach seeks to address this by utilizing several models with complementary strengths in order to minimize dataset bias and enhance cross-dataset performance. While the results indicate that generalization remains a challenge, the ensemble framework enhances stability and robustness across different datasets, demonstrating its potential for real-world no-reference IQA applications. Furthermore, mixed-dataset training can be used as a potential strategy to improve generalization by exposing the model to a more diverse range of distortions and content variations during training. This approach may help mitigate dataset-specific biases and further enhance cross-dataset performance.
\subsection*{Ablation Study}
The ablation study provides critical insights into the contributions and significance of each component in the proposed framework, as detailed in Table \ref{tab:ablation}. The removal of the meta-learner caused a notable performance degradation, with PLCC decreasing from 0.8840 to 0.8156, underscoring its vital role in effectively integrating predictions from base models. This performance drop highlights the efficiency of the meta-learner in leveraging the complementary strengths of the base models, optimizing their collective output, and justifying the computational cost of 12.5 GPU hours for the full framework compared to 8.3 GPU hours when using a simpler averaging strategy.\\
The quality-aware loss function demonstrated its importance by significantly outperforming the MS) loss function. Replacing the custom loss with MSE resulted in a PLCC reduction to 0.7925, indicating its inadequacy in capturing the nuanced relationships required for accurate perceptual quality assessment. This finding affirms that the quality-aware loss function's design aligns closely with human perceptual judgments, achieving superior accuracy for a moderate computational cost of 11.0 GPU hours.\\
Fine-tuning the model using the proposed pre-training strategy further illustrated its advantages over a generic ImageNet-pretrained model. The former not only achieved faster convergence but also delivered higher final accuracy, with a PLCC improvement from 0.7683 to 0.8840. This demonstrates the efficacy of pre-training on a quality-aware dataset in providing task-specific feature representations, which justify the computational investment.\\
The iterative selection of base learners through a SLR model further highlights the thoughtful design of the framework. Each base model contributes uniquely to the final prediction, and their inclusion is determined by their collective synergy rather than individual predictive performance. The removal of any base learner, as shown in the table, resulted in noticeable performance drops, with computational costs for individual base models ranging from 9.8 to 10.2 GPU hours.
\begin{table*}[!htbp]
\centering
\caption{Results of ablation study with computational cost analysis}
\label{tab:ablation}
\resizebox{\textwidth}{!}{
\begin{tabular}{lccc}
\toprule
\textbf{Experiment} &
\textbf{PLCC} &
\textbf{SROCC} &
\textbf{Computational Cost (GPU Hours)} \\
\midrule
Original Model                          & 0.8840 & 0.8765 & 12.5 \\
Meta-Learner Removal (Simple Averaging) & 0.8156 & 0.7892 & 8.3  \\
Removing Base Model 1                   & 0.8421 & 0.8173 & 10.2 \\
Removing Base Model 2                   & 0.8354 & 0.8117 & 10.1 \\
Removing Base Model 3                   & 0.8298 & 0.8059 & 9.8  \\
Removing Base Model 4                   & 0.8332 & 0.8104 & 9.9  \\
Using MSE Loss Function                 & 0.7925 & 0.7647 & 11.0 \\
Using ImageNet Pretrained Model         & 0.7683 & 0.7402 & 10.5 \\
\bottomrule
\end{tabular}
}
\end{table*}

\subsection*{Computational Runtime Analysis}
In order to assess the computational performance of the MetaQAP, computational runtime analysis is conducted. Inference time for various standard resolution images was measured over multiple iterations, and the average time for SD (480p), HD (720p), and Full HD (1080p) image sizes is reported for different models in Table \ref{tab:run-time}. The results depict that although our ensemble model generates better perceptual quality predictions, it increases computational cost relative to individual models.\\
Potential optimizations to address increased computational complexity for resource constrained settings are knowledge distillation, quantization, and model pruning without significantly compromising performance. Furthermore, efficient architectures such as MobileNet-V2 and EfficEfficientNet-B0 could be incorporated into the ensemble to increase efficiency. Future research will investigate these strategies to improve real-time applicability.\\
Although the ensemble model is still computationally expensive, by using the recommended optimizations one can strike a balance between accuracy and efficiency, so rendering the approach more feasible for practical uses.
\begin{table}[!ht]
\centering
\caption{Runtime Performance (Inference Time for various models in milliseconds)}
\begin{tabular}{llll}
\toprule
\multicolumn{1}{|c|}{\textbf{Model}} & \multicolumn{1}{c|}{\textbf{480p (SD)}} & \multicolumn{1}{c|}{\textbf{720p (HD)}} & \multicolumn{1}{c|}{\textbf{1080p (Full HD)}} \\ 
\midrule
ResNet50                             & $\sim$12 ms                                   & $\sim$25 ms                                   & $\sim$48 ms                                         \\ 
Inception-V3                         & $\sim$15 ms                                   & $\sim$30 ms                                   & $\sim$55 ms                                         \\ 
DenseNet-201                         & $\sim$20 ms                                   & $\sim$42 ms                                   & $\sim$78 ms                                         \\ 
NasNet-Large                         & $\sim$35 ms                                   & $\sim$68 ms                                   & $\sim$130 ms                                        \\ 
Ensemble Model                       & $\sim$50 ms                                   & $\sim$95 ms                                   & $\sim$180 ms                                        \\ 
\bottomrule
\end{tabular}
\label{tab:run-time}
\end{table}
\subsection*{Error Analysis}
While the proposed MetaQAP model demonstrates strong overall performance, error analysis provides valuable insights into the strengths and limitations of the proposed IQA model when applied to authentically distorted images. Unlike artificial distortions, authentic ones involve complex, natural degradation patterns with varying intensities, making quality prediction more challenging. Analyzing high and low prediction accuracy helps identify patterns and improvement areas.
\subsubsection*{Strong Prediction Cases}
\begin{figure}
    \centering
    \includegraphics[width=1\linewidth]{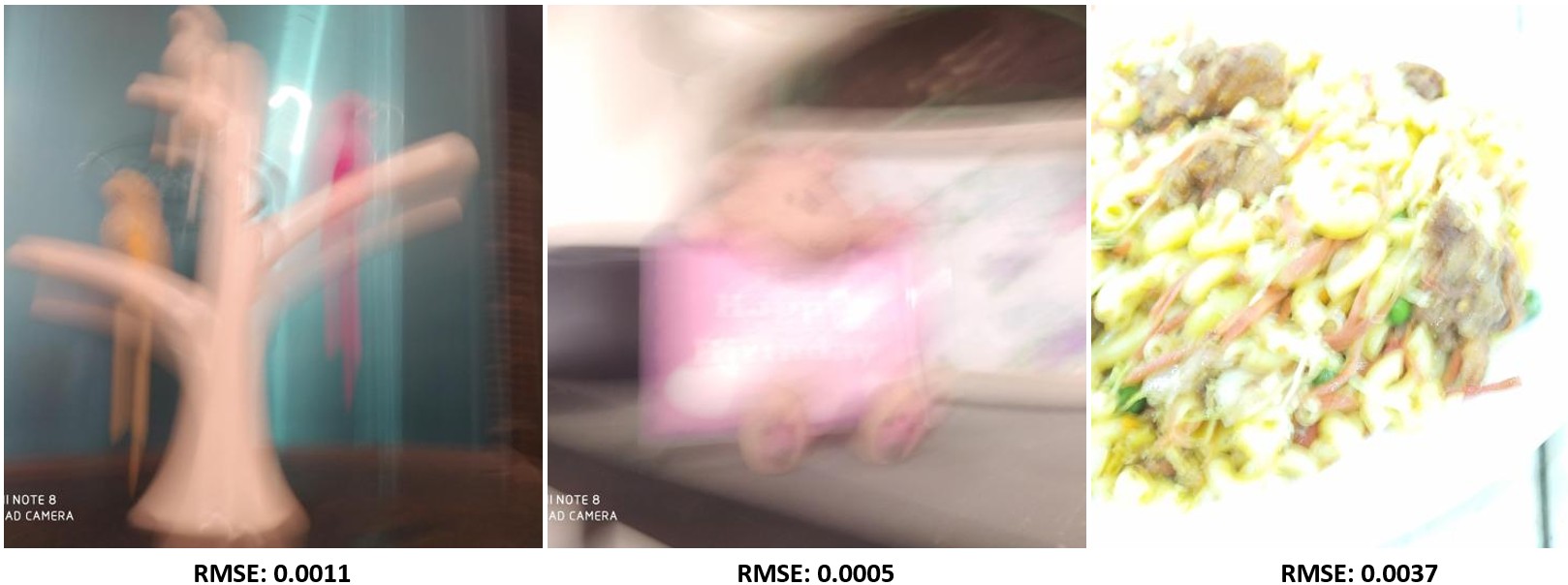}
    \caption{Sample cases of BIQ2021 dataset images \cite{ahmed2022biq2021} with low prediction error across various distortions i.e. noise (high ISO),  motion blur, and overexposure.}
    \label{fig:lowError}
\end{figure}

Figure \ref{fig:lowError} highlights three representative cases where the proposed model achieved exceptionally low prediction error, demonstrating its effectiveness in handling specific types of distortions. The first image, degraded by noise due to a high ISO setting, is accurately assessed by the model, yielding a minimal RMSE of 0.0081. Similarly, the second image, affected by motion blur, is evaluated with near-perfect quality prediction, achieving an RMSE of just 0.0005. Furthermore, the third image, which is overexposed, represents another scenario where the model excels, with a prediction error of 0.0011 RMSE.\\
These cases illustrate the model's ability to generalize well for distortions that exhibit consistent and straightforward patterns, such as noise, motion blur, and overexposure. Additionally, the model demonstrates strong performance on images with compression artifacts, further emphasizing its capability to capture regular degradation patterns effectively.

\subsubsection*{Challenging Prediction Cases}
\begin{figure}
    \centering
    \includegraphics[width=1\linewidth]{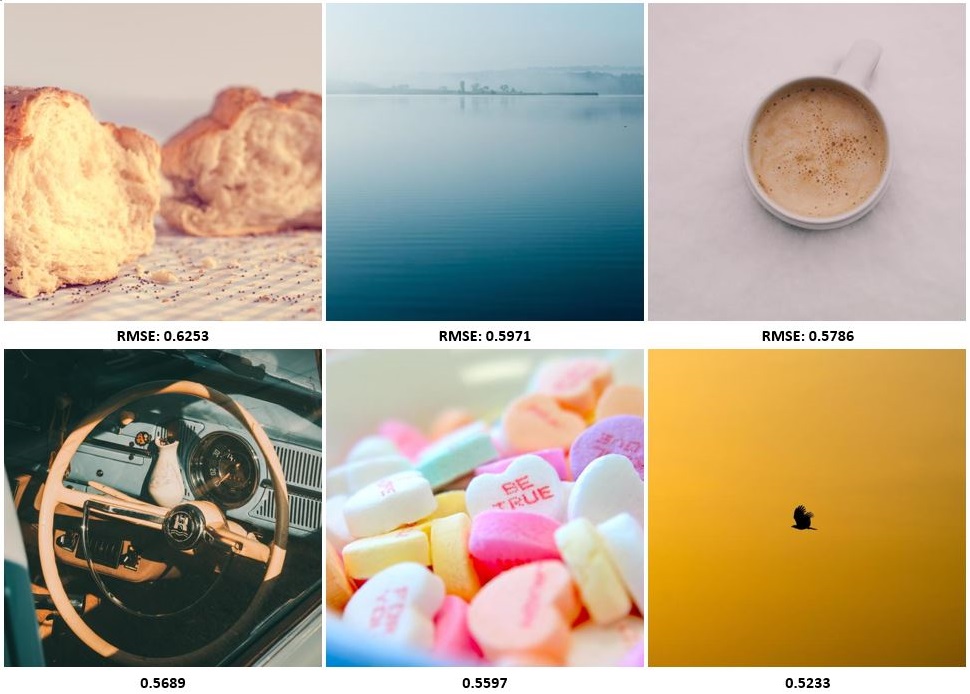}
    \caption{Sample cases of BIQ2021 dataset images \cite{ahmed2022biq2021} with high prediction error across various challenging scenarios such as regions with varying background and foreground quality, unusual angle, non-uniform illumination, mixed-quality regions and sparse details.}
    \label{fig:highError}
\end{figure}

Figure \ref{fig:highError} presents six challenging scenarios where the model exhibited poor performance, resulting in higher prediction errors. The first image highlights a disparity in quality between foreground and background objects, leading to a significant prediction error of 0.6253. Similarly, the second scenario depicts an image affected by haze, where varying levels of detail across different regions contribute to an error of 0.5971.\\
Another challenging case, shown in the third image, involves an object captured at an unusual angle, obscuring key quality aspects and causing the model to misinterpret the overall quality, leading to an error of 0.5786. In the fourth scenario, non-uniform illumination poses a significant challenge; regions exposed to sunlight appear overexposed, while others remain dark, resulting in a prediction error of 0.5689.\\
The fifth scenario features an image with multiple small, vibrant objects and varying foreground and background quality, making it difficult for the model to assess overall quality accurately. Lastly, the sixth image presents a scenario where a flying object is captured against a relatively plain sky. Due to poor lighting, the object appears as a silhouette, leading to a prediction error of 0.5233.
\subsubsection*{Cross-Dataset Variability Cases}
Cross-dataset evaluation highlights the model's ability to generalize across diverse datasets with varying distortion patterns and characteristics. However, certain cases reveal performance gaps due to the inherent differences between training and testing datasets.\\
When the model trained on KonIQ-10K was evaluated on LiveCD, it generally performed well but faced challenges with images exhibiting extreme sensor noise. These noise patterns, often linked to specific camera hardware, are underrepresented in the KonIQ-10K dataset. For example, a particularly noisy image from LiveCD resulted in a relatively high RMSE of 0.6508, reflecting the model's difficulty in adapting to such distortions.\\
The model trained on LiveCD struggled when tested on BIQ2021, particularly with high-contrast color saturation distortions. These distortions, characterized by vivid hues and exaggerated color contrasts, are less common in LiveCD. A vivid landscape image from BIQ2021, featuring over-saturated greens and blues, led to an RMSE of 0.5123, highlighting the model’s limitations in handling such scenarios.
\subsubsection*{Recommendations for Improvement}
The analysis of scenarios with low prediction error, high prediction error, and cross-dataset variability highlights areas where the proposed model can be refined to further enhance its performance and generalizability. Based on the observations, the following recommendations are proposed:
\begin{itemize}
    \item \textbf{Enhance Dataset Diversity:}\\
    To address scenarios where the model struggles, such as images with mixed-focus regions, non-uniform illumination, or high-contrast color saturation, it is crucial to augment the training dataset with more examples that capture these challenging conditions. Incorporating images with naturally occurring distortions, such as haze, chromatic aberrations, and mixed-quality regions, will enable the model to learn robust representations of these complex patterns.
    \item \textbf{Refine Pre-Training Strategy:}\\
    While the quality-aware pre-training strategy has proven effective, expanding the pre-training dataset to include a broader range of authentic distortions can improve the model's ability to generalize to unseen datasets. Incorporating real-world examples of sensor noise, motion blur, and overexposure would enhance the model’s capacity to handle cross-dataset variability.
    \item \textbf{Incorporate Attention Mechanisms:}\\
    Scenarios involving disparity in quality between the foreground and background objects, or regions with non-uniform illumination, suggest that the model may benefit from incorporating spatial attention mechanisms. These mechanisms can help the model focus on relevant regions of the image, improving its ability to make quality predictions based on the most perceptually significant areas.
    \item \textbf{Leverage Adaptive Loss Functions:}\\
    The current quality-aware loss function effectively balances error reduction and correlation maximization, but further enhancements could be explored. For example, adaptive loss functions that assign higher weights to underrepresented distortions or challenging cases (e.g., non-uniform illumination or haze) could help the model prioritize difficult scenarios during training.
    \item \textbf{Conduct Domain-Specific Fine-Tuning:}\\
    To minimize cross-dataset variability issues, domain-specific fine-tuning should be explored. For example, models trained on datasets like KonIQ-10K could be fine-tuned on LiveCD or BIQ2021 to better capture dataset-specific distortion patterns, such as sensor noise or high-contrast color saturation.   
\end{itemize}
\subsection*{Discussion}
Assessing image quality remains a multifaceted and subjective challenge due to the diverse factors influencing human perception, such as image content, distortion type, and intensity. These complexities are further exacerbated in no-reference image quality assessment \cite{ahmed2021piqi,ravela2019no}, where the absence of a reference image makes accurate evaluation particularly challenging. This problem is highly relevant in real-world applications like image sharing, video streaming, and automated quality control systems, where images often undergo uncontrolled and varied degradation during acquisition, storage, compression, or transmission. Despite the progress in no-reference approaches, traditional methods primarily rely on datasets with artificially generated distortions, which fail to capture the complexity and variability of authentic distortions present in real-world scenarios \cite{ahmed2022biq2021}.\\
Our research aims to bridge this gap by focusing exclusively on authentically distorted datasets, addressing the need for models that can generalize effectively to real-world conditions. To this end, we proposed a comprehensive framework that incorporates several novel contributions. First, we employed a pre-training strategy on a quality-aware dataset to provide the model with a strong foundation for recognizing diverse distortion patterns \cite{hendrycks2019using,aslam2023vrl}. This was followed by fine-tuning the model on authentically distorted datasets, enhancing its ability to align with human perceptual judgments \cite{ahmed2022biq2021,hosu2020koniq}. Additionally, the introduction of a quality-aware loss function, which optimizes both error minimization and perceptual correlation, significantly improved predictive accuracy. Finally, a meta-learner was integrated to combine predictions from multiple base models, leveraging their complementary strengths to enhance overall performance \cite{zhu2020metaiqa,wei2022perceptual}.\\
The effectiveness of the proposed framework was demonstrated through extensive experiments. Baseline evaluations of individual CNN models provided insights into their capacity to handle specific distortions, while the meta-learner substantially improved performance by optimally aggregating these predictions. Evaluation on benchmark datasets revealed that our approach consistently outperformed existing methods, achieving higher PLCC and SROCC values across all datasets \cite{ghadiyaram2015massive,hosu2020koniq,ahmed2022biq2021}. Cross-dataset evaluations further highlighted the model's generalization capabilities, although certain challenges were identified, such as handling sensor noise and high-contrast color saturation, which were underrepresented in the training datasets. These observations underscore the importance of training data diversity to improve robustness.\\
The ablation study reinforced the critical role of each component in the framework. Removing the meta-learner or replacing the quality-aware loss function with a simpler alternative, such as MSE, led to substantial performance degradation, highlighting the necessity of these components. Similarly, fine-tuning the model using the proposed pre-training strategy yielded superior results compared to using a generic ImageNet-pretrained model, further validating the importance of task-specific pre-training \cite{hosu2020koniq,ahmed2022deep,aslam2023vrl}.\\
Error analysis provided additional insights into the model's performance. The model excelled in scenarios involving consistent distortions, such as Gaussian blur and compression artifacts, which were well-represented in the training data. However, it struggled with more complex cases, such as mixed-focus regions (e.g., sharp foregrounds with blurred backgrounds) and non-uniform illumination. These findings suggest the need for incorporating spatial attention mechanisms or region-based evaluation strategies to address such challenges \cite{hosu2020koniq,ahmed2022biq2021}.\\
These findings suggest that the proposed framework offers significant advancements in no-reference IQA for authentically distorted images. By addressing key limitations of traditional methods and incorporating innovative components, our approach delivers a robust and generalizable solution for real-world quality assessment tasks. Future work will focus on further improving the model’s generalization capabilities by integrating domain-specific fine-tuning, adaptive loss functions, and attention-based learning to address the challenges identified in this study.
\section{Conclusion}
This study introduces a novel approach for no-reference image quality assessment by leveraging quality-aware pre-training and meta-learning to significantly enhance predictive performance. Through extensive experimental evaluations, the proposed MetaQAP model consistently outperforms existing IQA methods across multiple benchmark datasets, showcasing superior accuracy, robustness, and generalization capabilities. Key contributions of this research include the development of a quality-aware loss function tailored for perceptual alignment, the pre-training of CNN models on quality-aware datasets to improve feature extraction, and the integration of a meta-learner to create an optimized ensemble model. The ablation study highlights the importance of each component within the model architecture. The removal of the meta-learner or the substitution of the quality-aware loss function with a simpler alternative results in significant performance degradation, underscoring their critical roles in the overall framework. Additionally, the cross-dataset evaluation demonstrates the model's strong generalizability, successfully addressing the challenges of diverse and complex distortion patterns encountered in realistic scenarios. By addressing the complexities of authentic distortions, the proposed MetaQAP framework provides a robust and practical solution for real-world IQA applications. This research not only establishes a strong foundation for tackling the intricacies of no-reference IQA but also opens avenues for future exploration. Advancements in loss functions, domain-specific training strategies, and ensemble learning techniques hold the potential to further refine no-reference IQA models, paving the way for continued progress in the field.
\section*{Acknowledgement}
The authors extend their appreciation to the Deanship of Research and Graduate Studies at King Khalid University for funding this work through Large Group Project under grant number (RGP.2/556/45). Princess Nourah bint Abdulrahman University Researchers Supporting Project number (PNURSP2026R333), Princess Nourah bint Abdulrahman University, Riyadh, Saudi Arabia.
\section*{Declarations}
\begin{itemize}
\item \textbf{Funding:} The funding for the study is provided Deanship of Research and Graduate Studies at King Khalid University through Large Group Project under grant number (RGP.2/556/45) and  Princess Nourah bint Abdulrahman University Researchers Supporting Project number (PNURSP2026R333).
\item \textbf{Conflict of Interest:} The authors declare no conflicts of interest. 
\item \textbf{Data Availability:} The dataset used in this study are available at:\\
LiveCD: https://live.ece.utexas.edu/research/ChallengeDB/index.html\\
KonIQ-10K: https://database.mmsp-kn.de/koniq-10k-database.html\\
BIQ2021: https://www.kaggle.com/datasets/nisarahmedrana/biq2021
\item \textbf{Code Availability:} The code for the model and experiments is accessible via GitHub at: https://github.com/nisarahmedrana/MetaQAP. 
\item \textbf{Author Contributions:} N.A. performed implementation, experiments, analysis of results and writeup of the manuscript. G.S. assisted with dataset preparation, critical evaluation and pre-processing. N. Alturki contributed to the development of the quality-aware loss function and supervised the pre-training experiments. N. Alasbali supported the implementation and fine-tuning of the CNN models. All authors reviewed and approved the final manuscript.
\end{itemize}
\bibliography{sn-bibliography}

\end{document}